\documentclass[10pt,twocolumn,letterpaper]{article}

\usepackage{iccv}
\usepackage{times}
\usepackage{epsfig}
\usepackage{graphicx}
\usepackage{amsmath}
\usepackage{amssymb}
\usepackage[T1]{fontenc}
\usepackage[utf8]{inputenc}
\usepackage{authblk}
\usepackage[breaklinks=true,bookmarks=false]{hyperref}
\usepackage[capitalize]{cleveref}
\crefname{section}{Sec.}{Secs.}
\Crefname{section}{Section}{Sections}
\Crefname{table}{Table}{Tables}
\crefname{table}{Tab.}{Tabs.}
\usepackage[accsupp]{axessibility}
\usepackage{multicol}
\usepackage{multirow}
\usepackage[table]{xcolor}

\iccvfinalcopy 

\def\iccvPaperID{****} 

\ificcvfinal\fi
\begin{document}

\title{Fast Adversarial Training with Smooth Convergence}

\author{Mengnan Zhao}
\author{Lihe Zhang\thanks{Corresponding author}}
\author{Yuqiu Kong}
\author{Baocai Yin}
\affil{Dalian University of Technology, China \\{\tt\small zmnwelcome@mail.dlut.edu.cn, \{zhanglihe, yqkong, ybc\}@dlut.edu.cn}}

\renewcommand\Authands{ and }

\maketitle
\ificcvfinal\thispagestyle{empty}\fi

\begin{abstract}
	Fast adversarial training (FAT) is beneficial for improving the adversarial robustness of neural networks.
However, previous FAT work has encountered a significant issue known as catastrophic overfitting when dealing with large perturbation budgets, \ie the adversarial robustness of models declines to near zero during training. 
	To address this, we analyze the training process of prior FAT work and observe that catastrophic overfitting is accompanied by the appearance of loss convergence outliers.
	Therefore, we argue a moderately smooth loss convergence process will be a stable FAT process that solves catastrophic overfitting.
    To obtain a smooth loss convergence process, we propose a novel oscillatory constraint (dubbed ConvergeSmooth) to limit the loss difference between adjacent epochs. The convergence stride of ConvergeSmooth is introduced to balance convergence and smoothing.
Likewise, we design weight centralization without introducing additional hyperparameters other than the loss balance coefficient.
	Our proposed methods are attack-agnostic and thus can improve the training stability of various FAT techniques.
	Extensive experiments on popular datasets show that the proposed methods efficiently avoid catastrophic overfitting and outperform all previous FAT methods. 
Code is available at \url{https://github.com/FAT-CS/ConvergeSmooth}.
\end{abstract}

\section{Introduction}

Recent breakthroughs in deep learning \cite{krizhevsky2017imagenet, shankar2020evaluating} have aroused researchers' interest in the security of neural networks \cite{xu2020adversarial, zhang2022towards, wei2022cross}.
In particular, the advanced research proves the vulnerability of deep models to adversarial attacks \cite{dong2018boosting, goodfellow2014explaining, papernot2016limitations}. For instance, tiny crafted perturbations can fool models in various fields into making wrong decisions \cite{he2019parametric, zheng2021mc, sriramanan2020guided, li2020yet}.
Considering the security risks brought by adversarial attacks \cite{jia2020fooling, xu2019exact, sharma2018attend, xie2021enabling}, there is a quickly growing body of work \cite{xie2017adversarial, xu2019exact, jia2020fooling} on improving the adversarial robustness of neural networks. 
Among them, adversarial training is widely applied by practitioners \cite{shafahi2019adversarial, guo2018long}.

In recent years, projected gradient descent based adversarial training (PGD-AT) \cite{madry2017towards, song2018improving} has been widely employed for its stability and effectiveness. 
However, this mechanism is computationally expensive. 
It requires multiple gradient descent steps to generate the adversarial training data \cite{wang2019bilateral}. 
An alternative of PGD-AT is the fast adversarial training (FAT) \cite{park2021reliably}, which only adopts a single-step fast gradient sign method (FGSM) \cite{goodfellow2014explaining} to generate training data. 
Compared to PGD-AT, FAT can efficiently train models, but easily falls into catastrophic overfitting \cite{Wong2020, rice2020overfitting}.

A number of FAT methods have been proposed to mitigate catastrophic overfitting.
For example, Wong et al. \cite{Wong2020} use randomly initialized perturbations to enhance the diversity of adversarial perturbations.
Based on it, Andriushchenko et al. \cite{Andriushchenko2020} raise a complementary regularizer named GradAlign to maximize the gradient alignment between benign and adversarial examples explicitly. 
Similarly, NuAT \cite{sriramanan2021towards} and FGSM-MEP \cite{Jiag2022prior} adopt nuclear norm or p-norm to regularize the adversarial training, thereby increasing the prediction alignment between benign and adversarial examples.
However, the above methods can only resolve catastrophic overfitting within the limited perturbation budget ($\xi \le$ 8/255).
$\xi$ specifies the perturbation degree of adversarial training data generated by various attacks.
Besides, models trained by small perturbations are vulnerable to adversarial attacks with a large $\xi$, \eg the models trained by NuAT and FGSM-MEP at $\xi$ = 8/255 perform 53\% and 54\% robustness against the PGD-50 attack with $\xi$ = 8/255, but only 22\% and 20\% robustness against the same attack with $\xi$ = 16/255 respectively.
Therefore, we aspire to prevent catastrophic overfitting to improve the adversarial robustness of neural models at larger perturbation budgets.

\begin{figure*}[t]
    \begin{center}
        
        \includegraphics[width=1\linewidth]{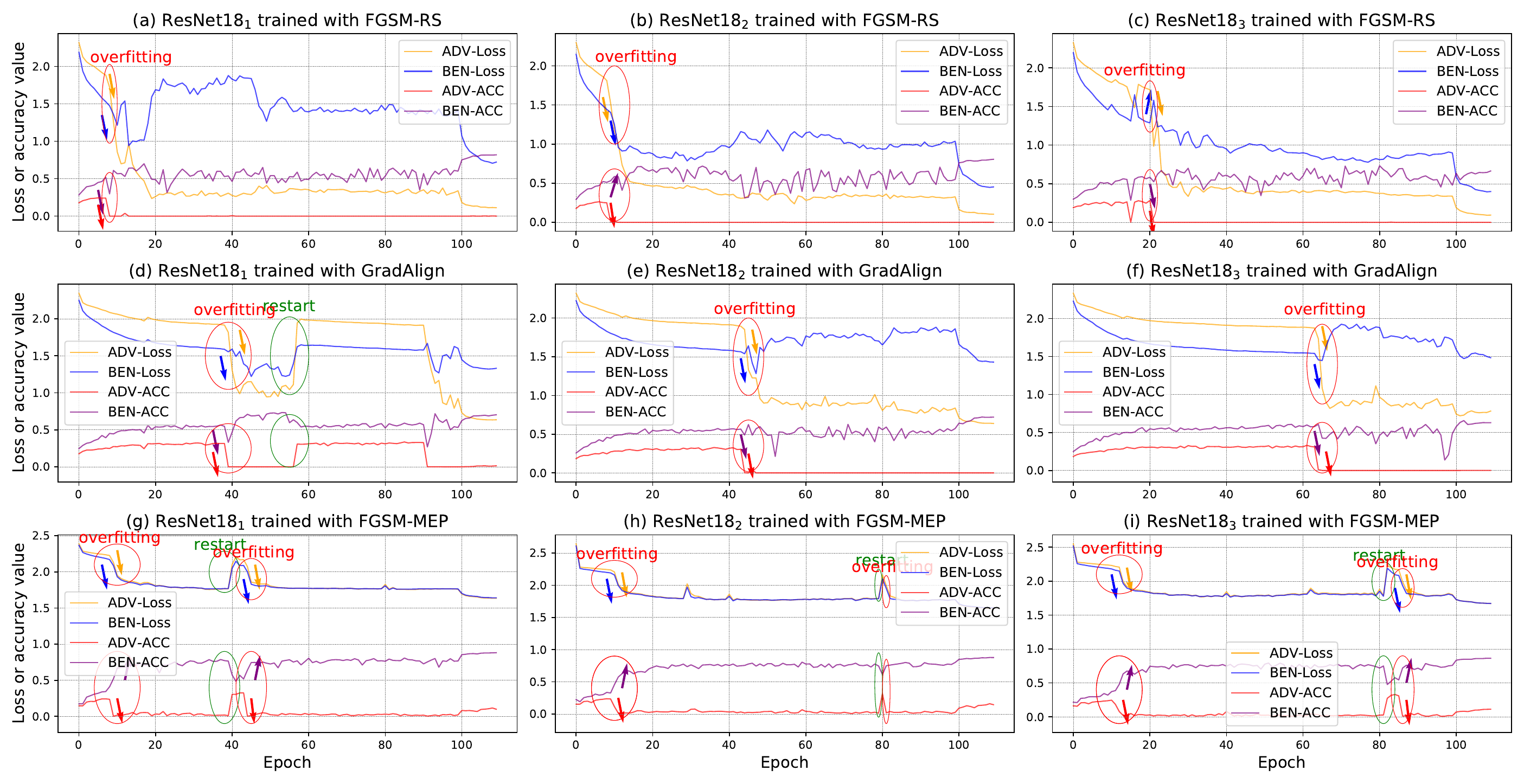}
    \end{center}
    \vspace{-3mm}
    \caption{The training process of previous FAT methods with ResNet18 on CIFAR-10. Each FAT method is trained 3 times. $\xi$ = 16/255. ADV-Loss and BEN-Loss denote the classification loss of models to adversarial and benign examples during training, respectively. ADV-Acc and BEN-Acc represent the classification accuracy of models to adversarial and benign examples during testing, respectively. }
    \label{fig1}
    \vspace{-2mm}
\end{figure*}

By analyzing adversarial training processes of representative work, we observe that catastrophic overfitting is usually accompanied by a slight fluctuation in the classification loss for benign samples and a sharp drop in the classification loss for adversarial examples.
This motivates us to question whether a smooth loss convergence process is also a stable FAT process.
Moreover, we find that an oscillating adversarial training phase may restart the FAT process after catastrophic overfitting.
Fig. \ref{fig1} shows the details.

According to these observations, we introduce an oscillatory constraint that limits the difference in loss between adjacent training epochs, called ConvergeSmooth.
A dynamic convergence stride of ConvergeSmooth is designed considering the nonlinear decay rate of loss functions.
Inspired by the smoothness of loss convergence, we further verify the effect of the proposed weight centralization on model stability. Weight centralization refers to taking the weights average of the previously trained models as the convergence center of the current model weight.
Our proposed methods are attack-agnostic and thus can be combined with existing adversarial strategies in FAT, such as FGSM-RS and FGSM-MEP, to evaluate their performance.

The contributions are summarized in four aspects:
{\bf (1)} We verify that previous FAT works still suffer from catastrophic overfitting at a large $\xi$ and then study catastrophic overfitting from the perspective of convergence instability of loss functions;
{\bf (2)} We propose a smooth convergence constraint, ConvergeSmooth, and design a dynamic convergence stride for it, to help various FAT methods avoid catastrophic overfitting on different perturbation budgets;
{\bf (3)} The weight centralization is proposed without introducing extra hyperparameters other than the loss balance coefficient to stabilize FAT;
{\bf (4)} Extensive experiments show that the proposed methods outperform the state-of-the-art FAT techniques in terms of efficiency, robustness, and stability.

\section{Related Work}

{\bf Adversarial attacks}.
Adversarial attacks are usually used to deceive deep-learning models.
Goodfellow et al. \cite{goodfellow2014explaining} first discuss the adversarial attack (FGSM) within the classification task.
They prove that adversarial examples $x^\prime$ generated by a single-step gradient backward can misclassify the model $f(\cdot; \mathbf{\theta})$ with high confidence. 
$\theta$ denotes the fixed model weights.
$x^\prime$ is generated by
\begin{equation}
    x^\prime = x + \xi\cdot \text{sgn}(\nabla_{x}\mathcal{L}(f(x; \mathbf{\theta}), y)), 
\end{equation}
where $x$ is an input image, $\xi$ represents the perturbation budget, sgn($\cdot$) means the sign function, $\mathcal{L}(\cdot)$ is usually the cross-entropy loss, $\nabla_{x}\mathcal{L}(\cdot)$ calculates the gradient of loss at $x$, and $y$ denotes the ground truth labels of $x$.
Following the FGSM \cite{goodfellow2014explaining}, researchers propose a series of attack methods based on the iterative gradient backward, \eg I-FGSM \cite{Kurakin2017}, MIM \cite{dong2018boosting}, and PGD \cite{madry2017towards}. 
Taking the PGD attack as an example, the adversarial example $x^\prime_{t+1}$ produced in the iteration $t$+1 can be formulated as
\begin{equation}
    x^\prime_{t+1} = \text{clip}_\xi(x^\prime_{t}+ \epsilon\cdot \text{sgn}(\nabla_{x^\prime_{t}}\mathcal{L}(f(x^\prime_{t}; \mathbf{\theta}), y))), 
\end{equation}
where $\epsilon$ denotes the single-step stride and $\text{clip}_\xi$ refers to projecting adversarial perturbations to a $\xi$-ball.

{\bf Adversarial training}. 
Madry et al. \cite{madry2017towards} formalize the adversarial training as a min-max optimization problem,
\begin{equation}\label{eq1}
    \min_\theta \mathbb{E}_{(x, y)\sim D}[\max_{\delta\in[-\xi, \xi]} \mathcal{L}(f(x^\prime; \theta), y) ],
\end{equation}
where $x^\prime$ = $x$ + $\delta$, $\delta$ represents the adversarial perturbations generated by various attacks such as PGD and FGSM. $D$ is the data generator.
The internal maximization maximizes the classification loss to generate adversarial perturbations with fixed model weights. 
The external minimization minimizes the classification loss on the generated adversarial examples when optimizing the model weights.

Actually, there is a trade-off between computational efficiency and adversarial robustness in recent adversarial training methods.
Compared to PGD-AT \cite{madry2017towards}, FGSM-based fast adversarial training (FAT) accelerates the training process but damages the robustness of models for the problem of catastrophic overfitting \cite{kim2021understanding, jia2022boosting}.
To mitigate this issue, 
Wong et.al \cite{Wong2020} demonstrate that the FGSM with a random start strategy (FGSM-RS) can achieve comparable performance against the PGD-AT.
ZeroGrad \cite{Golgooni2021} zeroes the elements of gradient that are too small to craft the perturbations.
enforces the model loss to increase with the increase in perturbation size.
Besides, GradAlign \cite{Andriushchenko2020} prevents catastrophic overfitting by maximizing the alignment between gradients of benign samples and adversarial examples.

Similarly, NuAT \cite{sriramanan2021towards} introduces a nuclear norm regularization between logits for benign and adversarial examples and uses the Bernoulli noise as the initial perturbation.
ATAS \cite{huang2022fast} learns an instantiated adaptive step size that is inversely proportional to the gradient norm. It applies the adversarial perturbations from the previous epoch as the initialization of FGSM in the current training phase.
In addition, Jia et al. \cite{Jiag2022prior} propose several prior-guided initialization methods to replace the random start strategy of FGSM-RS. 
Specifically,
FGSM-BP adopts the adversarial perturbations from the previous batch as the attack initialization in the current batch.
FGSM-MEP employs a momentum mechanism to combine all adversarial perturbations from the previous epochs to yield adversarial initialization.

Although these FAT methods have resolved catastrophic overfitting at small perturbation budgets, they still suffer from catastrophic overfitting under larger perturbation budgets (\eg $\xi$ = 16/255).
Unlike these methods, we revisit the catastrophic overfitting problem from the perspective of loss convergence instability and prevent this exception by limiting the magnitude of loss fluctuations during training.

\section{Proposed Method}
In this section, we first study the performance of previous FAT methods when subjected to a large perturbation budget $\xi$. 
Then, the training processes of these methods have been analyzed for a better understanding of catastrophic overfitting.
Finally, we detail the proposed methods.

\subsection{Performance of FAT methods on a large $\xi$}
Previous FAT techniques avoid catastrophic overfitting at small $\xi$ ($\xi$ $\le$ 8/255).
Here, we investigate their performance on a large $\xi$ ($\xi$ = 16/255 by default).

{\bf FGSM-RS \cite{Wong2020}:} Based on Eq. (\ref{eq1}), this method adopts the samples with uniformly random perturbations $\delta_0\sim \mathcal{U}(-\xi, \xi)$ as the attack initialization of FGSM,
\begin{equation}\label{eq2}
	\begin{aligned}
		\min_\theta \mathbb{E}_{(x, y)\sim D}[\max_{{\delta_0 + \delta}\in[-\xi, \xi]} \mathcal{L}(f(x^\prime; \theta), y) ],\\
		x^\prime = x + \delta_0 + \delta \quad s.t.  \quad x^\prime\in[0, 1], \quad \delta_0\sim \mathcal{U}(-\xi, \xi).
	\end{aligned}
\end{equation}

{{\bf GradAlign \cite{Andriushchenko2020}:}}
This approach increases the gradient alignment between benign samples $x$ and perturbed samples $x + \delta_{0}$, which is denoted as
\begin{equation}\label{eq5}
\centering
    \begin{aligned}
        	\mathbb{E}_{(x, y)\sim D}[1 - \cos(\nabla_{x}\mathcal{L}(x, \theta),  \nabla_{x + \delta_{0}}\mathcal{L}(x + \delta_{0}, \theta)],\\
\mathcal{L}(x, \theta) = \mathcal{L}(f(x; \theta), y).
    \end{aligned}
\end{equation}
$\cos(\cdot)$ computes the cosine similarity between two matrices.

{ {\bf FGSM-MEP \cite{Jiag2022prior}:}}
Different from FGSM-RS, this method generates the initialization perturbations $\delta_0$ based on all historical adversarial perturbations from the previous epochs and introduces a regularization expressed as
\begin{equation}\label{eqq5}
\mathbb{E}_{(x, y)\sim D}[||f(x^\prime; \theta) - f(x + \delta_{0}; \theta)||_2^2],
\end{equation}
where $||\cdot||_2^2$ denotes the squared $\mathcal{L}_2$ distance.

Fig. \ref{fig1} shows their detailed adversarial training processes of ResNet18 \cite{he2016deep} on CIFAR-10 \cite{krizhevsky2009learning}. 
We find that they fall into catastrophic overfitting during the 5$\sim$20$_{\text{th}}$, 40$\sim$65$_{\text{th}}$ and 10$\sim$15$_{\text{th}}$ epochs, respectively. 
Graphical analysis of various models and datasets is given in the supplement.


\subsection{{Analysis of training process}}
From Fig. \ref{fig1}, we discover several typical phenomena of catastrophic overfitting:
1) Change from a smooth convergence state to an irregular fluctuation state;
2) A slight change (increase or decrease) in the standard classification loss $\mathcal{L}(x, \theta)$ and a rapid descent of the adversarial classification loss $\mathcal{L}(x^\prime, \theta)$;
3) Rapid decline in the classification accuracy of adversarial examples.

After catastrophic overfitting, the methods depicted in Fig. \ref{fig1} are capable of restarting a stable FAT process, even though catastrophic overfitting may occur again.
The observed phenomena during the FAT restart are:
1) Change from an irregular fluctuation state to a smooth convergence state;
2) Rapid increase in both $\mathcal{L}(x, \theta)$ and $\mathcal{L}(x^\prime, \theta)$ after a period of decline;
3) Rapid decrease in the classification accuracy of benign samples, while the classification accuracy of adversarial examples experiences a rapid increase.

On this basis, we can make the following conclusions.
1) $\mathcal{L}(x,\theta)$ is more stable than $\mathcal{L}(x^\prime, \theta)$ and models are prone to overfitting to adversarial perturbations;
2) Catastrophic overfitting is closely correlated to the convergence instability of adversarial training;
3) Exceptions in $\mathcal{L}(x, \theta)$ and $\mathcal{L}(x^\prime, \theta)$ occur simultaneously;
4) An oscillating adversarial training phase may trigger the FAT process to restart.

\subsection{Smooth convergence for the stable FAT}
Next, we describe the proposed method in detail.

{\bf Why did previous methods fail to prevent catastrophic overfitting?}
Despite the improvement in diversity achieved through random initialization in FGSM-RS, models are still susceptible to overfitting adversarial perturbations.
GradAlign and FGSM-MEP enhance the stability of adversarial training through the constraints in Eqs. (\ref{eq5}) and (\ref{eqq5}), respectively.
However, Eq. (\ref{eq5}) may reduce the stability of $\mathcal{L}(x, \theta)$.
Meanwhile, the prediction probability $f(x + \delta_{0}; \theta)$ in Eq. (\ref{eqq5}) is not the sweet spot to keep the FAT stable.
Unlike these methods, we ensure the convergence stability of both $\mathcal{L}(x, \theta)$ and $\mathcal{L}(x^\prime, \theta)$.

{\bf How can catastrophic overfitting be solved?}
Since the catastrophic overfitting is usually accompanied by a slight change in $\mathcal{L}(x, \theta)$ and a sharp decline in $\mathcal{L}(x^\prime, \theta)$, we consider a smooth loss convergence process to be a stable FAT process that resolves this issue.
To this end, a complementary constraint $\mathcal{L}_{CS}$  for Eq. (\ref{eq1}) is proposed,
\begin{equation}\label{eqc7}
    \min_{\theta_{t}}\mathbb{E}_{(x, y)\sim D}[\mathcal{L}(x_{t}^\prime, \theta_{t}) + \mathcal{L}_{CS}(t)],
\end{equation}
which can limit the difference in losses between adjacent epochs, expressed as
\begin{equation}\label{eq7} 
\begin{aligned}
    \mathcal{L}_{CS}(t) = w_1\cdot|\mathcal{L}(x_t^\prime, \theta_{t}) - \mathcal{L}(x_{t-1}^\prime, \theta_{t-1})| + \\
w_2\cdot|\mathcal{L}(x, \theta_{t}) - \mathcal{L}(x, \theta_{t-1})|, \quad s.t. \quad\mathcal{C}(x) = 1,
\end{aligned}
\end{equation}
where $\theta_{t}$ represents the model weights of the $t_{\text{th}}$ training epoch.
$w_1$ and $w_2$ are hyper-parameters, $w_1, w_2\in[0,1]$. $|\cdot|$ calculates the absolute value. 
$x_{t}^\prime$ = $x$ + $\delta_{0,t} + \delta_{t}$.
$\delta_{0,t}$ can be replaced by various attack initialization perturbations and $\delta_{t}$ is generated by 
\begin{equation}\label{eqv8}
\max_{\delta_{t}\in[-\xi, \xi]}\mathcal{L}(x_{t}^\prime, \theta_{t-1}),
\end{equation}
where $\theta_{t-1}$ is fixed in generating $\delta_{t}$.

In the practical implementation of Eq. (\ref{eq7}), storing and computing $\mathcal{L}(x_{t-1}^\prime, \theta_{t-1})$ and $\mathcal{L}(x, \theta_{t-1})$ can consume a significant amount of memory. To overcome this challenge, we introduce $u_{t-1}^\prime$ and $u_{t-1}$ to replace these terms, respectively. $u_{t-1}^\prime = \mathbb{E}_{(x, y)\sim D}\mathcal{L}(x_{t-1}^\prime, \theta_{t-1})$ and $u_{t-1} = \mathbb{E}_{(x, y)\sim D}\mathcal{L}(x, \theta_{t-1})$ denote the mathematical expectation of loss in the $t$-$1_\text{th}$ training epoch.

We observe that training instability is primarily caused by overfitting or underfitting of small amounts of data. Thus, in Eq. (\ref{eq7}), the additional loss term is only applied to partial data selected by the crafted condition $\mathcal{C}(\cdot)$.
Notably, exceptions in $\mathcal{L}(x, \theta)$ and $\mathcal{L}(x^\prime, \theta)$ occur simultaneously during training. 
Hence, $\mathcal{L}(x_t^\prime, \theta_{t})$ and $\mathcal{L}(x_t, \theta_{t})$ share the same condition.
$\mathcal{C}$ is constructed by the distance between the pointwise loss $\mathcal{L}(x, \theta_{t})$ and the mean value $u_{t-1}$.
This is because over-fitted or under-fitted data often yield excessively high or low classification losses.

\begin{equation} 		\label{eq12}
	\begin{aligned}
 \mathcal{C}(x) = (|\mathcal{L}(x, \theta_{t}) - u_{t-1}| \ge \gamma_t),
\end{aligned}
\end{equation}
where $\gamma_t$ ($\gamma_t$ $\geq$ 0) represents the convergence stride to select abnormal data.
It is crucial to choose an appropriate value for $\gamma_t$ to ensure effective training.
When $\gamma_t = 0$, the adversarial training process may fail to converge as it forces the predictions of all samples to remain unchanged.
On the other hand, catastrophic overfitting occurs when $\gamma_t = \infty$, leading to undesirable outcomes.
Considering that the loss difference $d_{t-1}$ ($d_{t-1}$ = $|u_{t-1} - u_{t-2}|$) between two adjacent epochs tends to decrease non-linearly, $\gamma_t$ should be a variable that varies during the training process.

\begin{equation} 		\label{eq13}
	\centering
		\gamma_t = \min(\max(d_{t-1}, \gamma_{min}), \gamma_{max}),
\end{equation}
where $\gamma_{min}$ and $\gamma_{max}$ are hyper-parameters.
$\gamma_{max}$ controls the maximum convergence speed. 
$\gamma_{min}$ ensures that the training process is not too smooth when $d_{t-1}\rightarrow$ 0.

{\bf Example-based ConvergeSmooth.} 
This approach adds the constraint individually to each sample, described as 
\begin{equation}
\begin{aligned}
    \mathcal{L}_{CS}^E(t) = w_1\cdot|\mathcal{L}(x_t^\prime, \theta_{t}) - u_{t-1}^\prime| + 
w_2\cdot|\mathcal{L}(x, \theta_{t}) - u_{t-1}|, \\ s.t. \quad |\mathcal{L}(x, \theta_{t}) - u_{t-1}| > \gamma_t.
\end{aligned}
\end{equation}

{\bf Batch-based ConvergeSmooth.} 
Likewise, we can apply the complementary constraint to a data batch, 
\begin{equation}
\begin{aligned}
    \mathcal{L}_{CS}^B(t) = w_1\cdot|u_B^\prime - u_{t-1}^\prime| + 
w_2\cdot|u_B - u_{t-1}|, \\ s.t. \quad |u_B - u_{t-1}| > \gamma_t,
\end{aligned}
\end{equation}
where $u_{B}^\prime = \mathbb{E}_{(x, y)\sim B}\mathcal{L}(x^\prime_t, \theta_t)$, $u_{B} = \mathbb{E}_{(x, y)\sim B}\mathcal{L}(x, \theta_t)$.

{\bf Weight centralization.}
To mitigate the problem of manual parameter tuning, we introduce the weight centralization without requiring extra hyperparameters other than the coefficient $w_3$, $|\theta_t - \theta_{t-1}| = 0\rightarrow |\mathcal{L}(x,\theta_t) - \mathcal{L}(x,\theta_{t-1})| = 0$,
\begin{equation}\label{eq16}
    \mathcal{L}_{CS}^W(t) = w_3\cdot ||\theta_t - \frac{1}{len(\phi)}\cdot\sum_{j\in\phi}\theta_j||_p,
\end{equation}
where $\|\cdot\|_p$ means the p-norm function ($p$ = 2), and $\phi$ denotes a set of previous model weights.
The model weights $\theta_t$ are restricted to the center $\frac{1}{len(\phi)}\cdot\sum_{j\in\phi}\theta_j$.
The reasons behind Eq. (\ref{eq16}) are:
1) The initial training process is stable, as indicated in Fig. \ref{fig1};
2) Models at different training epochs tend to have similar weights after convergence \cite{tolpegin2020data}.

\section{Experimental Results}
\subsection{Experimental settings}
{\bf Details.} To demonstrate the effectiveness of the proposed method,
we conduct comprehensive experiments on several benchmark datasets, \ie CIFAR-10 \cite{krizhevsky2009learning}, CIFAR-100 \cite{krizhevsky2009learning}, and Tiny ImageNet \cite{deng2009imagenet}. 

Following previous works \cite{Wong2020, li2020towards, zhang2022revisiting, Jiag2022prior}, we adopt ResNet18 \cite{he2016deep} as the backbone on the CIFAR-10 and CIFAR-100, and choose PreActResNet18 \cite{he2016identity} on the Tiny ImageNet.
In all experiments, models are optimized using the SGD optimizer with a batch size of 128, weight decay of 5e-4, and momentum of 0.9.
The initial learning rates on CIFAR-10, CIFAR-100, and Tiny ImageNet are set as 0.1, 0.1, and 0.01, respectively.
Then, we optimize models with a total training epoch of 110 and decay the learning rate at the 100$_{\text{th}}$ and 105$_{\text{th}}$ epoch with a factor of 0.1.

We apply the proposed ConvergeSmooth in combination with two attack initialization methods on CIFAR-10 and CIFAR-100, FGSM-RS \cite{Wong2020} and FGSM-MEP \cite{Jiag2022prior}. Additionally, for the Tiny ImageNet, we use FGSM-BP \cite{Jiag2022prior} as the initialization method. As mentioned in \cite{Jiag2022prior}, FGSM-MEP requires consuming memory to store the previous adversarial perturbations, which limits its application on large datasets. Regarding the hyperparameters, we set $\gamma_{max}$ in Eq. (\ref{eq13}) as 0.03, 0.06, and 0.03 for CIFAR-10, CIFAR-100, and Tiny ImageNet, respectively. $\gamma_{max}$ = 1.5 $\cdot$ $\gamma_{min}$ and $\xi$ = 16/255. Specific details of hyperparameter settings for $w_{1\sim 3}$ are given in the supplement. All experiments are conducted on a single GeForce RTX 3090 GPU. 

{\bf Baselines.} 
We include advanced FAT methods as baselines, namely FGSM-RS \cite{Wong2020}, GradAlign \cite{Andriushchenko2020}, ZeroGrad \cite{Golgooni2021}, NuAT \cite{sriramanan2021towards}, ATAS \cite{huang2022fast}, and FGSM-MEP \cite{Jiag2022prior}.  

\begin{table}
    \tabcolsep = 0.05cm
    \begin{center}
    \begin{tabular}{cc|cccc}
    \hline\rowcolor{gray!25}
          CIFAR10&$\xi$&FGSM-RS& FGSM-MEP&B-RS&B-MEP  \\
         \hline
         SQ$\uparrow$&\multirow{2}{*}{16/255}&19.74&22.15&\underline{28.83} &{\bf 29.97}\\ 
          SQ+RayS$\uparrow$&&18.72&20.96&\underline{27.08}&{\bf 28.45}\\ 
         \hline
    \end{tabular}
    \end{center}
	\caption{ Quantitative results of various FAT methods against the Square and Ray-S attacks on CIFAR-10 with the backbone ResNet18. The number in bold and \underline{*} indicate the best and second-best results, respectively.}
    \label{b:my_label1}
\end{table}

\begin{table}
    \tabcolsep = 0.06cm
    \begin{center}
    \begin{tabular}{c|ccccccc}
    \hline\rowcolor{gray!25}
          CIFAR10&Clean& $\frac{8}{255}$$^\sharp$&$\frac{10}{255}$$^\sharp$&$\frac{12}{255}$$^\sharp$&$\frac{16}{255}$$^\sharp$&SQ+RayS &Time \\
         \hline
OAAT\cite{addepalli2022scaling}&71.30&44.29&38.56&33.68&28.48&21.06&183\\
AWP\cite{wu2020adversarial}&78.86&35.91&32.70&29.89&26.95&19.36&134\\
ATES\cite{Sitawarin2020Improving}&74.52&43.61&38.94&35.31&30.89&22.16&-\\
ExAT\cite{shaeiri2020towards}&80.78&49.21&42.15&36.98&31.26&23.08&70\\
E-MEP&69.84&{\bf 53.54}&{\bf 49.26}&{\bf 45.89}&{\bf 40.78}&\underline{23.78}&104\\
B-MEP&63.84&\underline{50.31}&\underline{46.77}&\underline{43.97}&\underline{40.13}&{\bf 28.45}&101\\
         \hline
    \end{tabular}
    \end{center}
    \caption{ Quantitative results of methods against various levels of perturbation with ResNet18 as the backbone and CIFAR-10 as the dataset. $\sharp$ denotes the PGD-10 attack.}
    \label{b:my_label2}
    \vspace{-4mm}
\end{table}
{\bf Evaluation metrics.}
We follow \cite{Wong2020, huang2022fast} and adopt FGSM \cite{goodfellow2014explaining}, PGD-10 \cite{madry2017towards}, PGD-20 \cite{madry2017towards}, PGD-50 \cite{madry2017towards}, C$\&$W \cite{carlini2017towards}, APGD-CE \cite{Francesco2020} and Autoattack (AA) \cite{Francesco2020} to evaluate the adversarial robustness of models.
PGD-n represents the PGD attack with n iterations. 
Autoattack combines APGD-CE and APGD-T (targeted APGD) as well as two complementary attacks (FAB \cite{5d1dce4b3a55ac56ce82a597} and Square attack \cite{Andriushchenko20202}). 

The training easily exhibits fluctuations, and thus  individual results can not objectively reflect the performance of methods.
For each method, we repeat the training process three times and report the average evaluation results of the best model across the three runs (\textit{${\text{mbest}}$}). Additionally, the supplement includes the best results (\textit{${\text{best}}$}) and the average evaluation of the final model from the three runs  (\textit{${\text{mfinal}}$}).

In the following section, `W-*', `E-*', and `B-*' mean `weight centralization', `example-based ConvergeSmooth', and `batch-based ConvergeSmooth', respectively.


\begin{figure}[t]
    \begin{center}
        
        \includegraphics[width=0.95\linewidth]{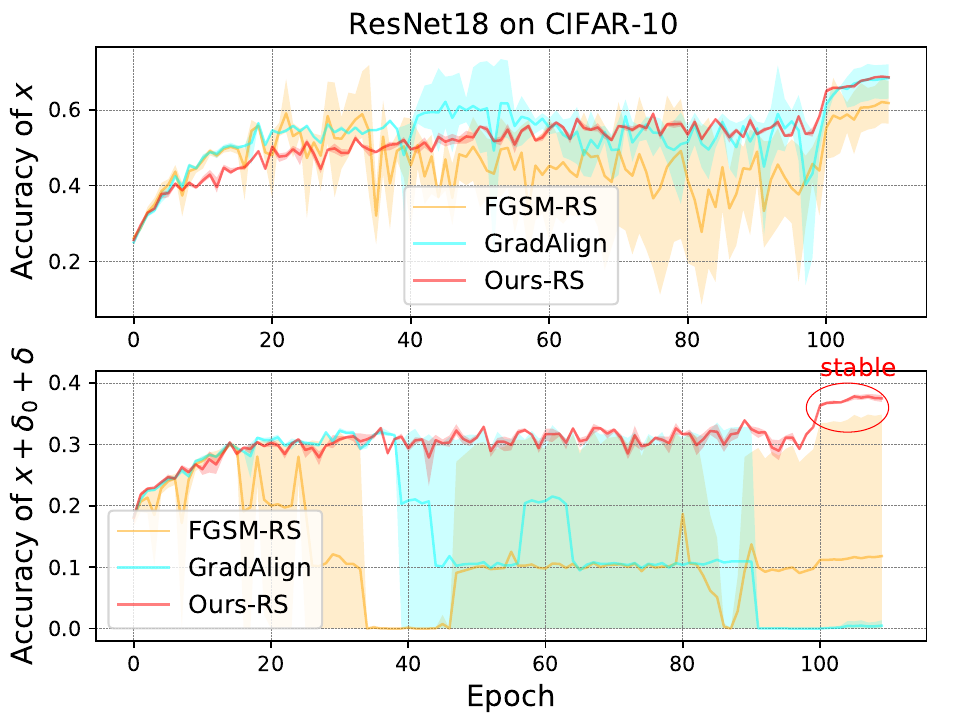}
    \end{center}
    \vspace{-4mm}
    \caption{  
    The training process of various FAT methods (PGD-10, $\xi$ = 16/255) on CIFAR-10.
    }
    \label{fig3}
	\vspace{-4mm}
\end{figure}

\begin{table*}
	\tabcolsep = 0.1cm
 \begin{center}
	\begin{tabular}{ccccc ccccc c} 
		\hline \rowcolor{gray!25}
		\multicolumn{2}{c}{Methods }    &  Clean$\uparrow$  &FGSM$\uparrow$ & PGD-10$\uparrow$  &  PGD-20$\uparrow$ & PGD-50$\uparrow$  & C$\&$W$\uparrow$  & APGD-CE$\uparrow$ & AA $\uparrow$ & Time (min)$\downarrow$ \\ 
		\hline
		\multicolumn{2}{c}{PGD-AT \cite{rice2020overfitting}}
		&65.30&46.31&40.73&35.08&33.92&30.84&33.08&26.29&{370}\\
		\hline\hline
		\multicolumn{2}{c}{FGSM-RS \cite{Wong2020}}
		&50.12&38.28&26.13&21.55&20.43&18.96&19.40&14.84& {67}\\
		\multicolumn{2}{c}{GradAlign \cite{Andriushchenko2020}}
		&58.17&39.87&33.12&26.81&24.99&22.63&23.98&17.02& {135}\\
		\multicolumn{2}{c}{ZeroGrad \cite{Golgooni2021}}
		&74.16&43.96&32.67&21.98&18.37&20.76&16.44&12.07& {67}\\

		\multirow{3}{*}{Ours}&{W-RS}
		&70.66&45.51&36.50&27.51&24.75&23.97&23.38&17.14& {71}\\

&{E-RS}
		&62.38&42.07&36.78&{\bf 30.80}&{\bf 28.90}&23.64&{\bf 27.71}&17.55&{77}\\

		&{B-RS}
		&65.42&{\bf 45.94}&{\bf 37.54}&30.01&27.85&{\bf 26.28}&26.52&{\bf 19.43}& {75}\\
		\hline
		\multicolumn{2}{c}{NuAT \cite{sriramanan2021towards}}
		&74.62	&44.92	&35.22	&25.93	&23.67	&24.07	&22.37	&18.43& {101}\\
		\multicolumn{2}{c}{ATAS$^*$ \cite{huang2022fast}}&64.11&-&31.39&-&28.15&-&-&21.09&-\\
		\hline
		\multicolumn{2}{c}{FGSM-MEP \cite{Jiag2022prior}}
		&53.32&36.24&31.85&27.28&26.56&22.10&26.08&18.98& {92}\\
		\multirow{2}{*}{Ours}&{E-MEP}
   &69.84&{\bf 47.18}&{\bf 40.90}&34.17&32.72&22.69&31.12&17.74& {104}\\
		&{B-MEP}
		&63.84&45.48&40.13&{\bf 34.21}&{\bf 32.95}&{\bf 28.19}&{\bf 32.04}&{\bf 23.68}&{101}\\
		\hline
	\end{tabular}
 \end{center}
	\caption{Quantitative results of various methods ($\xi$ = 16/255) on the CIFAR-10 with ResNet18 as the backbone.
    `ATAS$^*$' is the result of ATAS in \cite{huang2022fast}, which is superior to our reproduction.
    We train each method three times.
    The results represent the evaluation average between the best models of three training processes.
    Weight centralization and regularization in MEP do not work together.
	}
	\label{tab1}
\end{table*}

\begin{table*}
	\centering
	\tabcolsep = 0.09cm
 \begin{center}
	\begin{tabular}{ccccc ccccc c} 
		\hline \rowcolor{gray!25}
		Budgets ($\xi$)& Methods     &  Clean$\uparrow$  &FGSM$\uparrow$ & PGD-10$\uparrow$  &  PGD-20$\uparrow$ & PGD-50$\uparrow$  & C$\&$W$\uparrow$  & APGD-CE$\uparrow$ & AA $\uparrow$ & Time (min)$\downarrow$ \\ 
		\hline  
        \multirow{4}{*}{12/255}&
		{GradAlign \cite{Andriushchenko2020}}&66.52&43.06&31.66&27.95&26.04&27.05&25.99&21.65&135\\
		&NuAT \cite{sriramanan2021towards} &72.79&51.80&41.75&38.60&37.54&35.99&36.71&32.01&101\\
		&FGSM-MEP \cite{Jiag2022prior} &{\bf 74.71}&52.05&38.45&36.35&35.52&33.05&33.4&27.23&92\\
		&B-MEP (Ours)&72.63&{\bf 54.40}&{\bf 45.23}&{\bf 42.85}&{\bf 42.14}&{\bf 36.81}&{\bf 41.62}&{\bf 33.26}&{101}\\
		\hline
        \multirow{4}{*}{10/255}&
		{GradAlign \cite{Andriushchenko2020}}&83.10&55.23&36.58&30.47&28.64&31.01&26.51&23.84&135\\
		&NuAT \cite{sriramanan2021towards} &75.82&55.94&45.54&43.92&43.42&{\bf 41.39}&42.91&38.85&101\\
		&FGSM-MEP \cite{Jiag2022prior} &{\bf 83.43}&{\bf 59.51}&42.76&39.28&37.33&37.26&35.91&32.52&92\\
		&B-MEP (Ours)&75.96&57.26&{\bf 47.25}&{\bf 45.98}&{\bf 45.66}&{41.00}&{\bf 45.26}&{\bf 39.20}&{101}\\
  \hline
	\end{tabular}
 \end{center}
	\caption{ Quantitative results of FAT methods on various  $\xi$ with ResNet18 as the backbone and CIFAR-10 as the dataset. Models are trained and evaluated under the same $\xi$. }
	\label{tab34}
\end{table*}

\begin{table}
	\tabcolsep = 0.1cm
 \begin{center}
	\begin{tabular}{cccccc} 
		\hline \rowcolor{gray!25}
		$\xi$ = 16/255 &\multicolumn{2}{c}{CIFAR10}&\multicolumn{2}{c}{CIFAR100}&Time\\ 
            \cline{2-5}
            Methods& Clean & AA & Clean & AA &hours\\
		\hline
            GradAlign \cite{Andriushchenko2020}&55.58&13.45&35.93&7.13&8.79\\
            NuAT \cite{sriramanan2021towards}&74.25&13.19&20.48&7.30&7.66\\
            FGSM-MEP \cite{Jiag2022prior}&65.56&15.43&20.69&6.73&6.15\\
            B-MEP (Ours) &{\bf 69.94}&{\bf 24.27}&{\bf 48.64}&{\bf 12.05}&{ 6.72}\\
            \hline
	\end{tabular}
 \end{center}
	\caption{The adversarial accuracy of various FAT methods with WideResNet34-10 as the backbone. }
	\label{tabps}
 \vspace{-4mm}
\end{table}

\subsection{Significance of the Results}
Before evaluating the adversarial robustness of trained models, we demonstrate the significance of the results \cite{addepalli2022scaling}. Specifically, we show that adversarial attacks with $\xi$ = 16/255 rarely change the true label of input.
1) We generate adversarial examples for the FGSM-RS, FGSM-MEP, and our proposed methods on CIFAR10 and CIFAR100 datasets with $\xi$=16/255 using the non-targeted attacks such as FGSM, PGD, C$\&$W, APGD-CE, and the targeted attack APGD-T.
The majority of the generated examples retain their true labels;
2) Tab. \ref{b:my_label1} examines the robustness of models against an ensemble of the Square (SQ) \cite{Andriushchenko20202} and Ray-S \cite{chen2020rays} attacks, as these attacks generate strong oracle-invariant examples \cite{addepalli2022scaling};
3) We generate adversarial training data with $\xi$ = 16/255 and test the robustness on various levels of perturbation. 
Tab. \ref{b:my_label2} provide comparative experiments between our proposed method and other AT techniques, including OAAT \cite{addepalli2022scaling}, ExAT \cite{shaeiri2020towards}, ATES \cite{Sitawarin2020Improving} and AWP \cite{wu2020adversarial}. It is important to note that we re-implement these AT methods and apply them in the context of FAT.
Overall, our methods achieve optimal performance on both oracle-invariant attacks and classical evaluation attacks. Namely, the classical evaluation attacks used in $\xi$ = 8/255 are reliable and significant even when the value of $\xi$ is increased to 16/255.

Notably, our methods are plugged into FGSM-RS and FGSM-MEP,
which are fitted to the distribution of adversarial examples. 
Instead, methods such as OAAT are fitted to the distribution of benign samples and perform better on clean accuracy.
Therefore, we use the settings as in OAAT and then apply OAAT and our B-OAAT to the FAT task.
B-OAAT realizes 74.12\% (+2.82\% than OAAT) clean accuracy and 25.39\% (OAAT) on SQ+RayS.
\begin{table*}
	\tabcolsep = 0.09cm
 \begin{center}
	\begin{tabular}{cccccc ccccc } 
		\hline \rowcolor{gray!25}
		\multicolumn{2}{c}{Methods} & Clean$\uparrow$  &FGSM$\uparrow$ & PGD-10$\uparrow$  &  PGD-20$\uparrow$ & PGD-50$\uparrow$  & C$\&$W$\uparrow$  & APGD-CE$\uparrow$ & AA$\uparrow$ & Time (min)$\downarrow$ \\ 
		\hline
		\multicolumn{2}{c}{{PGD-AT}} 
		&40.57&24.72&21.46&17.94&17.38&14.98&17.02&12.63&{370}\\
		\hline\hline

		\multicolumn{2}{c}{{FGSM-RS \cite{Wong2020}}}
		&30.12&15.24&12.69&10.25&9.79&8.45&9.57&6.90&{67}\\
		\multicolumn{2}{c}{{GradAlign \cite{Andriushchenko2020}}}
		&31.91&15.71&12.56&10.28&9.71&8.22&9.47&6.61&{135}\\
		\multicolumn{2}{c}{{ZeroGrad \cite{Golgooni2021}}}
		&47.31&23.58&17.60&12.85&11.87&11.62&11.01&7.94&{67}\\
			\multirow{3}{*}{Ours}&{W-RS}
		&41.97&{\bf 26.30}&19.09&14.77&13.72&12.82&13.36&9.67&{72}\\

  &{E-RS}
  &41.09 & 22.33  & 18.78 & 15.19 &  {\bf 14.43} & 12.00 &{\bf 13.91}  &9.50&{78}\\
  &{B-RS}
		&41.47&25.98&{\bf 19.44}&{\bf 15.36}&14.34&{\bf 12.91}&13.71&{\bf 9.90}&{76}\\
  
		\hline
		\multicolumn{2}{c}{{NuAT \cite{sriramanan2021towards}}}
		&31.42&20.39&16.15&13.87&13.29&11.12&12.25&8.32&{101}\\
		\multicolumn{2}{c}{{ATAS \cite{huang2022fast}}}
		&55.63	&30.35	&15.31	&10.28	&8.49	&10.97	&6.30	&5.05&{70}\\
		\hline
		\multicolumn{2}{c}{{FGSM-MEP \cite{Jiag2022prior}}}
		&21.39&13.00&11.37&9.93&9.76&7.28&9.58&6.36&{92}\\
		\multirow{2}{*}{Ours}&{E-MEP}
		&44.00&24.46&20.59&17.11&16.59&13.37&16.05&10.97&{106}\\
		&{B-MEP}
		&41.86&{\bf 24.86}&{\bf 20.84}&{\bf 17.30}&{\bf 16.59}&{\bf 13.96}&{\bf 16.25}&{\bf 11.38}&{102}\\
		\hline
	\end{tabular}
 \end{center}
	\caption{Quantitative results of various methods ($\xi$ = 16/255) with ResNet18 as the backbone on CIFAR-100. 
	}
	\label{tab2}
\end{table*}
\begin{table*}[htpb]
	\centering
	\tabcolsep = 0.1cm
 \begin{center}
	\begin{tabular}{cccc ccccc cc} 
		\hline \rowcolor{gray!25}
		Methods    & Clean$\uparrow$  &FGSM$\uparrow$ & PGD-10$\uparrow$  &  PGD-20$\uparrow$ & PGD-50$\uparrow$  & C$\&$W$\uparrow$  & APGD-CE$\uparrow$ & AA$\uparrow$ & Time (hour)$\downarrow$ \\ 
		\hline
		{PGD-AT}
		&32.52&16.47&13.40&10.63&10.24&7.95&10.00&6.41&67.2\\
		\hline\hline
		{FGSM-RS \cite{Wong2020}}
		&27.48&12.46&9.59&6.97&6.47&4.99&6.19&3.63 &10.5\\
		{GradAlign \cite{Andriushchenko2020}}
		&28.65&13.80&10.40&7.78&7.08&5.75&6.59&3.90&20.9\\
		{ZeroGrad \cite{Golgooni2021}}
		&34.66	&12.26	&8.22	&5.29	&4.82	&3.71	&4.23	&2.21&10.5\\
		{NuAT \cite{sriramanan2021towards}}
		&35.24	&15.52	&12.07	&8.81	&8.13	&6.29	&7.68	&4.27 &24.6\\
		{FGSM-BP \cite{Jiag2022prior}}
		&20.02&10.18&8.38&6.98&6.77&4.56&6.57&3.62&14.3\\
		\hline
		{B-BP (Ours)}
		&33.51&{\bf 16.32}&{\bf 12.92}&{\bf 9.72}&{\bf 9.23}&{\bf 7.26}&{\bf 9.32}&{\bf 5.95}&{15.4}\\
		\hline
	\end{tabular}
 \end{center}
	\caption{Quantitative results of various methods ($\xi$ = 16/255) with PreActResNet18 as the backbone on Tiny ImageNet.
	}
	\label{tab3}
 \vspace{-2mm}
\end{table*}

\subsection{Results on CIFAR-10}
We conduct our initial experiments on CIFAR-10 using the ResNet18 backbone.
The default $\xi$ is set to 16/255.
Tab. \ref{tab1} presents the main comparisons.
The observations are as follows:
(1) Compared with previous FAT methods, the proposed approaches achieve optimal adversarial robustness against different attacks, \eg B-RS outperforms all RS-based methods and B-MEP is superior to all other methods.
Meanwhile, our methods exhibit similar performance to prior work in terms of clean classification accuracy.
(2) B-MEP realizes adversarial robustness approaching PGD-AT, \eg B-MEP performs 32.95\% robustness against the PGD-50 attack, only 0.97\% lower than PGD-AT;
(3) As for time consumption, B-RS takes less time (75 minutes) than GradAlign (135 minutes) and NuAT (101 minutes). 
This is because ConvergeSmooth only requires the additional regularization when $|u_B - u_{t-1}|$$>$$\gamma_t$ instead of adding constraints on all iterations.
In addition, B-MEP (102 minutes) takes a bit more calculation cost than FGSM-MEP (92 minutes) but much less than PGD-AT (370 minutes).
PGD-AT takes significantly longer than FAT works;
(4) B-RS and B-MEP successfully prevent catastrophic overfitting, $e.g.$ \textit{${\text{mfinal}}$} (in the supplement) only shows a slight reduction than \textit{${\text{mbest}}$} in terms of adversarial robustness.

Fig. \ref{fig3} visualizes the training process of various FAT approaches.
Compared with other methods, the proposed B-RS achieves optimal training stability and adversarial robustness, \eg the classification accuracy of our method fluctuates slightly during three adversarial training sessions.

{\bf Various budgets.} 
Similar experiments are performed for budgets 10/255 and 12/255.
Details are given in Tab. \ref{tab34}.
It is evident that the proposed method achieves optimal adversarial robustness across various perturbation budgets.

{\bf Various networks.} 
We then adopt WideResNet34 with a width factor of 10 \cite{zagoruyko2016wide} as the backbone.
The results are given in Tab. \ref{tabps}.
Our proposed B-MEP also prevents the wider architectures from catastrophic overfitting.

\subsection{Results on CIFAR-100}
The results on CIFAR-100 with the backbone ResNet-18 are presented in Tab. \ref{tab2}.
(1) FAT methods perform similar training consumption on CIFAR-10 and CIFAR-100 as  two datasets contain the same number and size of images;
(2) All proposed methods can realize stable adversarial training, as evidenced by the comparable results of \textit{${\text{mbest}}$} and \textit{${\text{mfinal}}$} (in the supplement);
(3) Among the RS-based FAT works, B-RS and E-RS achieve the best and second-best adversarial robustness against different attacks;
(4) B-MEP outperforms all other FAT methods;
(5) Although B-MEP performs slightly worse than PGD-AT, its training process is approximately 3.5 times faster than this competitor.
\begin{table*}
	\tabcolsep = 0.1cm
  \begin{center}
	\begin{tabular}{cccccccccc c} 
		\hline \rowcolor{gray!25}
		Dataset&$w_1$    & Clean$\uparrow$  &FGSM$\uparrow$ & PGD-10$\uparrow$  &  PGD-20$\uparrow$ & PGD-50$\uparrow$  & C$\&$W$\uparrow$  & APGD-CE$\uparrow$ & AA $\uparrow$ &Stability \\ 
		\hline
        \multirow{5}{*}{CIFAR100}&0.0&{\bf 48.13}&{\bf32.13}&{\bf24.25}&{\bf22.67}&{\bf22.21}&{\bf19.04}&{\bf 21.29}&{\bf 15.28}&$\star\star\star$\\
        &0.3&46.08&30.58&23.06&21.62&21.22&17.79&20.44&15.22&$\star\star\star$\\
        &0.5&44.04&29.96&22.64&21.34&20.99&17.55&19.16&13.84&$\star\star\star$\\
                &0.7&43.07&28.61&22.01&20.84&20.32&16.87&18.98&13.47&$\star\star\star$\\
                        &0.9&40.45&21.05&21.62&20.54&20.15&16.33&18.91&13.53&$\star\star\star$\\
        \hline
	\end{tabular}
 \end{center}
	\caption{Quantitative results of the proposed method on various  $w_1$ with ResNet18 as the backbone and the perturbation budget 12/255. `Stability' represents the number of times the model is stable in three training repetitions.  $\gamma_{max}$ and $w_2$ are set to 0.03 and 1, respectively.}
	\label{tab6}
\end{table*}

\begin{table*}
	\tabcolsep = 0.1cm
 \begin{center}
	\begin{tabular}{cccccccccc c} 
		\hline \rowcolor{gray!25}
		Dataset&$w_2$    & Clean$\uparrow$  &FGSM$\uparrow$ & PGD-10$\uparrow$  &  PGD-20$\uparrow$ & PGD-50$\uparrow$  & C$\&$W$\uparrow$  & APGD-CE$\uparrow$ & AA $\uparrow$ &Stability \\ 
		\hline
        \multirow{4}{*}{CIFAR100}
        &0.5&36.79&23.97&18.60&17.51&17.23&14.65&16.28&12.23&-\\
        &0.7&37.31&24.36&18.56&17.37&17.26&14.46&15.96&11.54&-\\
                &1.0&{\bf 48.13}&{\bf 32.13}&{\bf 24.25}&{\bf 22.67}&{\bf 22.21}&{\bf 19.04}&{\bf 21.29}&{\bf 15.28}&$\star\star\star$\\
                        &1.3&44.84&30.42&23.31&21.72&21.31&18.36&19.64&14.72&$\star\star\star$\\
        \hline
	\end{tabular}
 \end{center}
	\caption{Quantitative results on various  $w_2$ ($\gamma_{max}$ = 0.03 and $w_1$ = 0) with ResNet18 as the backbone and $\xi$ = 12/255. }
	\label{tab7}
\end{table*}

\subsection{Results on Tiny ImageNet}
We conduct experiments on the Tiny ImageNet using the PreActResNet18 backbone to demonstrate the scalability of the proposed method to large datasets.  The results are given in Tab. \ref{tab3}. B-BP achieves higher robustness among the FAT methods and comparable robustness to PGD-AT. As for training efficiency, B-BP (15.4 hours) requires slightly more computational cost than FGSM-BP (14.3 hours), but significantly less time than PGD-AT (67.2 hours).

\begin{figure}[t]
    \begin{center}
        \includegraphics[width=1\linewidth]{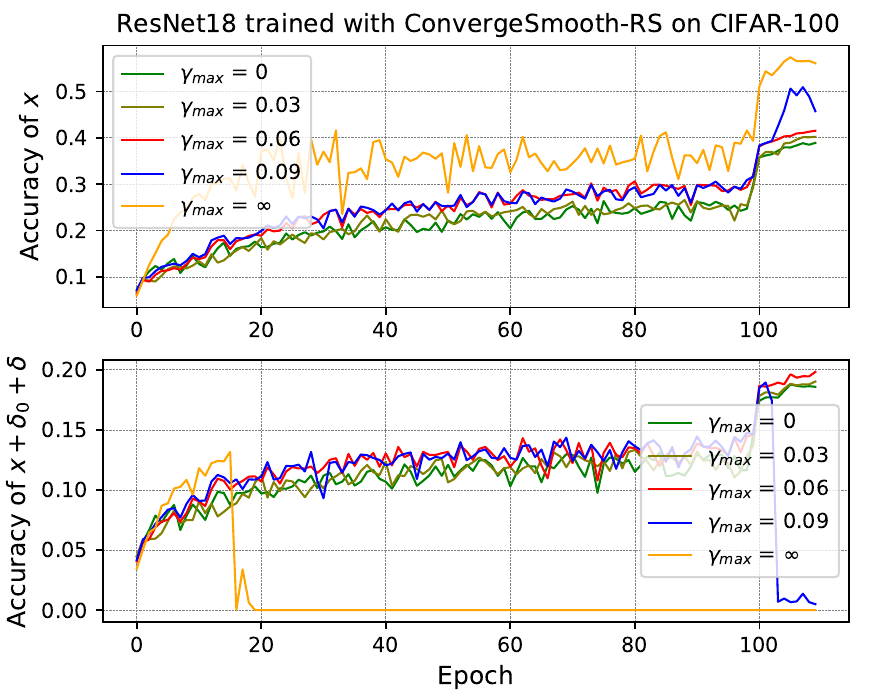}
    \end{center}
	\vspace{-3mm}
    \caption{    
Ablation study of $\gamma_{max}$ ($\gamma_{max}/\gamma_{min}$ = 1.5). We provide the classification accuracy of models (ResNet-18) to benign and adversarial examples (PGD-10, $\xi$ = 16/255) on CIFAR-100.
    }
    \label{fig4}
	\vspace{-4mm}
\end{figure}

\begin{figure}[t]
    \begin{center}
        \includegraphics[width=1\linewidth]{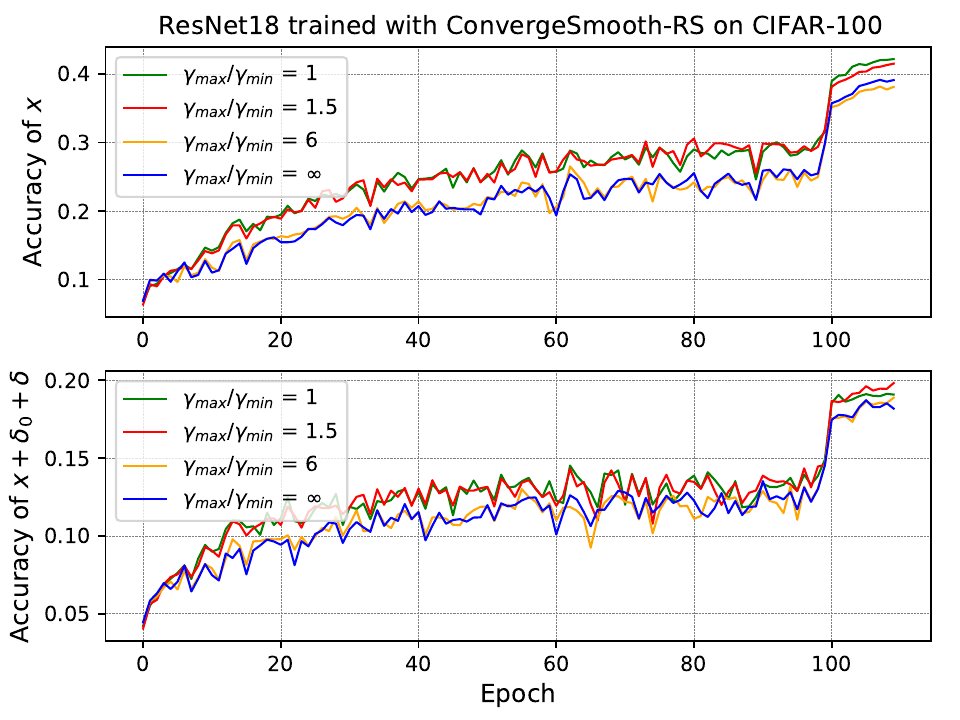}
    \end{center}
	\vspace{-2mm}
    \caption{
    Ablation study of $\gamma_{max}/\gamma_{min}$ ($\gamma_{max}$ = 0.06). We provide the classification accuracy of models (ResNet-18) to benign and adversarial examples (PGD-10, $\xi$ = 16/255) on CIFAR-100.
    }
    \label{fig5}
    \vspace{-2mm}
\end{figure}

\subsection{Ablation studies}
In the following, B-RS is selected as the backbone.
Figs. \ref{fig4} and \ref{fig5} conduct ablation studies on $\gamma_{max}$ and $\gamma_{max}/\gamma_{min}$ respectively.
In Fig. \ref{fig4}, the classification accuracy of both benign and adversarial examples increases with $\gamma_{max}$. 
However, the model trained with a large $\gamma_{max}$ ($>$0.09) suffers from catastrophic overfitting. 
Fig. \ref{fig5} shows that the model with $\gamma_{max}/\gamma_{min}$ = 1.5 achieves optimal robustness.
$\gamma_{max}/\gamma_{min}$ = 1 denotes the static convergence stride.
All models can be trained stably when $\gamma_{max}$ = 0.06.

Tabs. \ref{tab6} and \ref{tab7} provide the results of comparative experiments on $w_1$ and $w_2$ respectively.
We empirically set $w_2$ to 1 since too small $w_2$ cannot prevent overfitting and too large $w_2$ will affect the performance. 
In Tab. \ref{tab6}, $w_1$ = 0 achieves optimal adversarial robustness on CIFAR100 and all settings accomplish stable FAT.
According to Eqs. (\ref{eqc7}) and (\ref{eq7}), if $w_1$$\neq$0, for the data $x$ which satisfies $\mathcal{C}(x)$ = 1, $\mathcal{L}(x_{t}^\prime, \theta_{t})$ is assigned with higher weights (1+$w_1$) or lower weights (1-$w_1$), causing the model to excessively prioritize or neglect this portion of the data.
This may weaken the performance of models.
However, for Tiny ImageNet, models suffer catastrophic overfitting at $w_1$ = 0. Then, we gradually increase $w_1$  from 0.3 with a stride of 0.2 until the FAT process becomes stable.
In addition, our weight centralization method only introduces a balance coefficient $w_3$.
$w_3$ is set to 0.1 since catastrophic overfitting happens when $w_3$ = 0 and classification performance is degraded when $w_3$ = 0.2.
\section{Conclusion}
In this paper, we tackle the issue of catastrophic overfitting by focusing on the convergence stability of loss functions. Through experimental analysis, we find that this issue is accompanied by abnormal convergence behavior of losses. Motivated by this phenomenon, we propose complementary constraints, namely E-ConvergeSmooth and B-ConvergeSmooth, based on the adversarial training constraint. To alleviate the burden of parameter tuning, weight centralization is designed utilizing priors from previous epochs. Extensive experiments on different network architectures and datasets show that the proposed methods effectively solve catastrophic overfitting and exhibit superior robustness against various adversarial attacks.

{\bf Acknowledgments.} This work was supported by the National Key R\&D Program of China \#2018AAA0102000 and the National Natural Science Foundation of China \#62276046 and \#U19B2039.

{\small
\bibliographystyle{ieee_fullname}
\bibliography{egbib}

\begin{thebibliography}{10}\itemsep=-1pt

\bibitem{addepalli2022scaling}
Sravanti Addepalli, Samyak Jain, Gaurang Sriramanan, and R Venkatesh~Babu.
\newblock Scaling adversarial training to large perturbation bounds.
\newblock In {\em European Conference on Computer Vision}, pages 301--316.
  Springer, 2022.

\bibitem{Andriushchenko2020}
Flammarion~N Andriushchenko~M.
\newblock Understanding and improving fast adversarial training.
\newblock In {\em Advances in Neural Information Processing Systems}, pages
  16048--16059, 2020.

\bibitem{carlini2017towards}
Nicholas Carlini and David Wagner.
\newblock Towards evaluating the robustness of neural networks.
\newblock In {\em IEEE Symposium on Security and Privacy}, pages 39--57, 2017.

\bibitem{Sitawarin2020Improving}
Sitawarin Chawin, Chakraborty Supriyo, and Wagner David.
\newblock Improving adversarial robustness through progressive hardening.
\newblock {\em arXiv preprint arXiv:2003.09347}, 2020.

\bibitem{chen2020rays}
Jinghui Chen and Quanquan Gu.
\newblock Rays: A ray searching method for hard-label adversarial attack.
\newblock In {\em Proceedings of the 26th ACM SIGKDD International Conference
  on Knowledge Discovery \& Data Mining}, pages 1739--1747, 2020.

\bibitem{Francesco2020}
Francesco Croce and Matthias Hein.
\newblock Reliable evaluation of adversarial robustness with an ensemble of
  diverse parameter-free attacks.
\newblock In {\em International Conference on Machine Learning}, 2020.

\bibitem{deng2009imagenet}
Jia Deng, Wei Dong, Richard Socher, Li-Jia Li, Kai Li, and Li Fei-Fei.
\newblock Imagenet: A large-scale hierarchical image database.
\newblock In {\em 2009 IEEE conference on computer vision and pattern
  recognition}, pages 248--255. Ieee, 2009.

\bibitem{dong2018boosting}
Yinpeng Dong, Fangzhou Liao, Tianyu Pang, Hang Su, Jun Zhu, Xiaolin Hu, and
  Jianguo Li.
\newblock Boosting adversarial attacks with momentum.
\newblock In {\em Proceedings of the IEEE conference on computer vision and
  pattern recognition}, pages 9185--9193, 2018.

\bibitem{5d1dce4b3a55ac56ce82a597}
Croce Francesco and Hein Matthias.
\newblock Minimally distorted adversarial examples with a fast adaptive
  boundary attack.
\newblock In {\em International Conference on Machine Learning}, pages
  2196--2205. PMLR, 2020.

\bibitem{Golgooni2021}
Z. Golgooni, M. Saberi, M. Eskandar, and M.~H Rohban.
\newblock Zerograd: Mitigating and explaining catastrophic overfitting in fgsm
  adversarial training.
\newblock page arXiv preprint arXiv:2103.15476, 2021.

\bibitem{goodfellow2014explaining}
Ian~J Goodfellow, Jonathon Shlens, and Christian Szegedy.
\newblock Explaining and harnessing adversarial examples.
\newblock In {\em International Conference on Learning Representations (ICLR)},
  2015.

\bibitem{guo2018long}
Jiaxian Guo, Sidi Lu, Han Cai, Weinan Zhang, Yong Yu, and Jun Wang.
\newblock Long text generation via adversarial training with leaked
  information.
\newblock In {\em Proceedings of the AAAI conference on artificial
  intelligence}, 2018.

\bibitem{he2016deep}
Kaiming He, Xiangyu Zhang, Shaoqing Ren, and Jian Sun.
\newblock Deep residual learning for image recognition.
\newblock In {\em Proceedings of the IEEE conference on computer vision and
  pattern recognition}, pages 770--778, 2016.

\bibitem{he2016identity}
Kaiming He, Xiangyu Zhang, Shaoqing Ren, and Jian Sun.
\newblock Identity mappings in deep residual networks.
\newblock In {\em European conference on computer vision}, pages 630--645.
  Springer, 2016.

\bibitem{he2019parametric}
Zhezhi He, Adnan~Siraj Rakin, and Deliang Fan.
\newblock Parametric noise injection: Trainable randomness to improve deep
  neural network robustness against adversarial attack.
\newblock In {\em Proceedings of the IEEE/CVF Conference on Computer Vision and
  Pattern Recognition}, pages 588--597, 2019.

\bibitem{huang2022fast}
Zhichao Huang, Yanbo Fan, Chen Liu, Weizhong Zhang, Yong Zhang, Mathieu
  Salzmann, Sabine S{\"u}sstrunk, and Jue Wang.
\newblock Fast adversarial training with adaptive step size.
\newblock {\em arXiv preprint arXiv:2206.02417}, 2022.

\bibitem{jia2022boosting}
Xiaojun Jia, Yong Zhang, Baoyuan Wu, Jue Wang, and Xiaochun Cao.
\newblock Boosting fast adversarial training with learnable adversarial
  initialization.
\newblock {\em IEEE Transactions on Image Processing}, 31:4417--4430, 2022.

\bibitem{jia2020fooling}
Yunhan~Jia Jia, Yantao Lu, Junjie Shen, Qi~Alfred Chen, Hao Chen, Zhenyu Zhong,
  and Tao~Wei Wei.
\newblock Fooling detection alone is not enough: Adversarial attack against
  multiple object tracking.
\newblock In {\em International Conference on Learning Representations
  (ICLR'20)}, 2020.

\bibitem{kim2021understanding}
Hoki Kim, Woojin Lee, and Jaewook Lee.
\newblock Understanding catastrophic overfitting in single-step adversarial
  training.
\newblock In {\em Proceedings of the AAAI Conference on Artificial
  Intelligence}, volume~35, pages 8119--8127, 2021.

\bibitem{krizhevsky2009learning}
Alex Krizhevsky, Geoffrey Hinton, et~al.
\newblock Learning multiple layers of features from tiny images.
\newblock 2009.

\bibitem{krizhevsky2017imagenet}
Alex Krizhevsky, Ilya Sutskever, and Geoffrey~E Hinton.
\newblock Imagenet classification with deep convolutional neural networks.
\newblock {\em Communications of the ACM}, 60(6):84--90, 2017.

\bibitem{Kurakin2017}
A. Kurakin, I.J. Goodfellow, and S. Bengio.
\newblock Adversarial machine learning at scale.
\newblock In {\em International Conference on Learning Representations (ICLR)},
  2017.

\bibitem{li2020towards}
Bai Li, Shiqi Wang, Suman Jana, and Lawrence Carin.
\newblock Towards understanding fast adversarial training.
\newblock {\em arXiv preprint arXiv:2006.03089}, 2020.

\bibitem{li2020yet}
Qizhang Li, Yiwen Guo, and Hao Chen.
\newblock Yet another intermediate-level attack.
\newblock In {\em European Conference on Computer Vision}, pages 241--257.
  Springer, 2020.

\bibitem{madry2017towards}
Aleksander Madry, Aleksandar Makelov, Ludwig Schmidt, Dimitris Tsipras, and
  Adrian Vladu.
\newblock Towards deep learning models resistant to adversarial attacks.
\newblock In {\em International Conference on Learning Representations (ICLR)},
  2018.

\bibitem{Andriushchenko20202}
Andriushchenko Maksym, Croce Francesco, Flammarion Nicolas, and Hein Matthias.
\newblock Square attack: a query-efficient black-box adversarial attack via
  random search.
\newblock In {\em European Conference on Computer Vision}, 2020.

\bibitem{papernot2016limitations}
Nicolas Papernot, Patrick McDaniel, Somesh Jha, Matt Fredrikson, Z~Berkay
  Celik, and Ananthram Swami.
\newblock The limitations of deep learning in adversarial settings.
\newblock In {\em 2016 IEEE European symposium on security and privacy
  (EuroS\&P)}, pages 372--387. IEEE, 2016.

\bibitem{park2021reliably}
Geon~Yeong Park and Sang~Wan Lee.
\newblock Reliably fast adversarial training via latent adversarial
  perturbation.
\newblock In {\em Proceedings of the IEEE/CVF International Conference on
  Computer Vision}, pages 7758--7767, 2021.

\bibitem{rice2020overfitting}
Leslie Rice, Eric Wong, and Zico Kolter.
\newblock Overfitting in adversarially robust deep learning.
\newblock In {\em International Conference on Machine Learning}, pages
  8093--8104. PMLR, 2020.

\bibitem{shaeiri2020towards}
Amirreza Shaeiri, Rozhin Nobahari, and Mohammad~Hossein Rohban.
\newblock Towards deep learning models resistant to large perturbations.
\newblock {\em arXiv preprint arXiv:2003.13370}, 2020.

\bibitem{shafahi2019adversarial}
Ali Shafahi, Mahyar Najibi, Mohammad~Amin Ghiasi, Zheng Xu, John Dickerson,
  Christoph Studer, Larry~S Davis, Gavin Taylor, and Tom Goldstein.
\newblock Adversarial training for free!
\newblock volume~32, 2019.

\bibitem{shankar2020evaluating}
Vaishaal Shankar, Rebecca Roelofs, Horia Mania, Alex Fang, Benjamin Recht, and
  Ludwig Schmidt.
\newblock Evaluating machine accuracy on imagenet.
\newblock In {\em International Conference on Machine Learning}, pages
  8634--8644. PMLR, 2020.

\bibitem{sharma2018attend}
Vasu Sharma, Ankita Kalra, Sumedha~Chaudhary Vaibhav, Labhesh Patel, and
  Louis-Phillippe Morency.
\newblock Attend and attack: Attention guided adversarial attacks on visual
  question answering models.
\newblock In {\em Proc. 32nd Conf. Neural Inf. Process. Syst.(NeurIPS)}, pages
  1--6, 2018.

\bibitem{song2018improving}
Chuanbiao Song, Kun He, Liwei Wang, and John~E Hopcroft.
\newblock Improving the generalization of adversarial training with domain
  adaptation.
\newblock In {\em International Conference on Learning Representations (ICLR)},
  2019.

\bibitem{sriramanan2020guided}
Gaurang Sriramanan, Sravanti Addepalli, Arya Baburaj, et~al.
\newblock Guided adversarial attack for evaluating and enhancing adversarial
  defenses.
\newblock {\em Advances in Neural Information Processing Systems},
  33:20297--20308, 2020.

\bibitem{sriramanan2021towards}
Gaurang Sriramanan, Sravanti Addepalli, Arya Baburaj, et~al.
\newblock Towards efficient and effective adversarial training.
\newblock volume~34, pages 11821--11833, 2021.

\bibitem{tolpegin2020data}
Vale Tolpegin, Stacey Truex, Mehmet~Emre Gursoy, and Ling Liu.
\newblock Data poisoning attacks against federated learning systems.
\newblock In {\em Computer Security--ESORICS 2020: 25th European Symposium on
  Research in Computer Security, ESORICS 2020, Guildford, UK, September 14--18,
  2020, Proceedings, Part I 25}, pages 480--501. Springer, 2020.

\bibitem{wang2019bilateral}
Jianyu Wang and Haichao Zhang.
\newblock Bilateral adversarial training: Towards fast training of more robust
  models against adversarial attacks.
\newblock In {\em Proceedings of the IEEE/CVF International Conference on
  Computer Vision}, pages 6629--6638, 2019.

\bibitem{wei2022cross}
Zhipeng Wei, Jingjing Chen, Zuxuan Wu, and Yu-Gang Jiang.
\newblock Cross-modal transferable adversarial attacks from images to videos.
\newblock In {\em Proceedings of the IEEE/CVF Conference on Computer Vision and
  Pattern Recognition}, pages 15064--15073, 2022.

\bibitem{Wong2020}
Kolter J~Z. Wong~E, Rice~L.
\newblock Fast is better than free: Revisiting adversarial training.
\newblock In {\em International Conference on Learning Representations (ICLR)},
  2020.

\bibitem{wu2020adversarial}
Dongxian Wu, Shu-Tao Xia, and Yisen Wang.
\newblock Adversarial weight perturbation helps robust generalization.
\newblock {\em Advances in Neural Information Processing Systems},
  33:2958--2969, 2020.

\bibitem{Jiag2022prior}
Jia Xiaojun, Zhang Yong, Wei Xingxing, Wu Baoyuan, Ma Ke, Wang Jue, and Cao
  Xiaochun.
\newblock Prior-guided adversarial initialization for fast adversarial
  training.
\newblock In {\em Proceedings of the European conference on computer vision
  (ECCV)}, 2022.

\bibitem{xie2017adversarial}
Cihang Xie, Jianyu Wang, Zhishuai Zhang, Yuyin Zhou, Lingxi Xie, and Alan
  Yuille.
\newblock Adversarial examples for semantic segmentation and object detection.
\newblock In {\em Proceedings of the IEEE international conference on computer
  vision}, pages 1369--1378, 2017.

\bibitem{xie2021enabling}
Yi Xie, Zhuohang Li, Cong Shi, Jian Liu, Yingying Chen, and Bo Yuan.
\newblock Enabling fast and universal audio adversarial attack using generative
  model.
\newblock In {\em Proceedings of the AAAI Conference on Artificial
  Intelligence}, 2021.

\bibitem{xu2020adversarial}
Han Xu, Yao Ma, Hao-Chen Liu, Debayan Deb, Hui Liu, Ji-Liang Tang, and Anil~K
  Jain.
\newblock Adversarial attacks and defenses in images, graphs and text: A
  review.
\newblock {\em International Journal of Automation and Computing},
  17(2):151--178, 2020.

\bibitem{xu2019exact}
Yan Xu, Baoyuan Wu, Fumin Shen, Yanbo Fan, Yong Zhang, Heng~Tao Shen, and Wei
  Liu.
\newblock Exact adversarial attack to image captioning via structured output
  learning with latent variables.
\newblock In {\em Proceedings of the IEEE/CVF Conference on Computer Vision and
  Pattern Recognition}, pages 4135--4144, 2019.

\bibitem{zagoruyko2016wide}
Sergey Zagoruyko and Nikos Komodakis.
\newblock Wide residual networks.
\newblock {\em arXiv preprint arXiv:1605.07146}, 2016.

\bibitem{zhang2022towards}
Jie Zhang, Bo Li, Jianghe Xu, Shuang Wu, Shouhong Ding, Lei Zhang, and Chao Wu.
\newblock Towards efficient data free black-box adversarial attack.
\newblock In {\em Proceedings of the IEEE/CVF Conference on Computer Vision and
  Pattern Recognition}, pages 15115--15125, 2022.

\bibitem{zhang2022revisiting}
Yihua Zhang, Guanhua Zhang, Prashant Khanduri, Mingyi Hong, Shiyu Chang, and
  Sijia Liu.
\newblock Revisiting and advancing fast adversarial training through the lens
  of bi-level optimization.
\newblock In {\em International Conference on Machine Learning}, pages
  26693--26712. PMLR, 2022.

\bibitem{zheng2021mc}
Wenqiang Zheng and Yan-Fu Li.
\newblock Mc-fgsm: Black-box adversarial attack for deep learning system.
\newblock In {\em 2021 IEEE International Symposium on Software Reliability
  Engineering Workshops (ISSREW)}, pages 154--159. IEEE, 2021.

\end{thebibliography}
}

\end{document}